\documentclass{llncs}
\usepackage{graphicx}
\usepackage{booktabs}
\usepackage{marvosym}
\usepackage{multirow}
\usepackage[retainorgcmds]{IEEEtrantools}
\usepackage{amsmath,amssymb} 
\usepackage{color}
\usepackage{url}
\usepackage{amsfonts}
\usepackage{footnote}
\usepackage{xspace}

\usepackage{algorithmicx}
\usepackage{algorithm}
\usepackage{algpseudocode}

\usepackage{marvosym}

\usepackage[width=122mm,left=12mm,paperwidth=146mm,height=193mm,top=12mm,paperheight=217mm]{geometry}

\DeclareMathOperator*{\argmin}{\arg\!\min}
\def\etal{\emph{et al.}}

\graphicspath{{figures/}}

\usepackage{setspace}
\usepackage{mdwlist}
\usepackage{hyperref}

\newcommand{\hl}{\bgroup\markoverwith{\textcolor{yellow}{\rule[-.5ex]{2pt}{2.5ex}}}\ULon}

\renewcommand{\thepage}{
  \roman{page}
}

\begin{document}
\long\def\symbolfootnote[#1]#2{\begingroup%
\def\thefootnote{\fnsymbol{footnote}}\footnote[#1]{#2}\endgroup}

\newcommand{\logoEPFL}[0]{
  \begin{center}
    \includegraphics[scale=0.8]{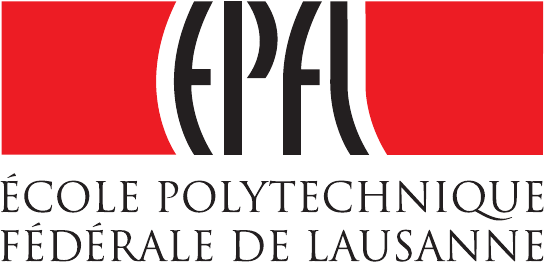}
  \end{center}
}

\renewcommand{\title}[1]{
  \vspace{1cm}
  \begin{center}
    \huge{#1}
  \end{center}
  \vspace{0.8cm}
}

\renewcommand{\author}[3]{
  \begin{center}
    \Large{#1} (\large{#2})\\
    \normalsize{\url{#3}}
  \end{center}
  \vspace{0.1cm}
}

\newcommand{\department}[1]{
  \begin{center}
    \large{#1}
  \end{center}
}

\newcommand{\reptype}[1]{
  \begin{center}
    \textbf{\large{#1}}
  \end{center}
}

\renewcommand{\date}[3]{
  \begin{center}
    \large{#1 #2, #3}
  \end{center}
}

\thispagestyle{empty}

\logoEPFL
\title{Beyond KernelBoost\symbolfootnote[1]{This work has been supported in part by the Swiss National Science Foundation.}}
\author{Roberto Rigamonti} {roberto.rigamonti@epfl.ch}
{http://cvlab.epfl.ch/~rigamont}
\author{Vincent Lepetit} {vincent.lepetit@epfl.ch}
{http://cvlab.epfl.ch/~lepetit}
\author{Pascal Fua} {pascal.fua@epfl.ch}
{http://cvlab.epfl.ch/~fua}
\vspace{0.4cm}
\department{School of Computer and Communication Sciences\\
  Swiss Federal Institute of Technology, Lausanne (EPFL)}

\vspace{1.0cm}
\reptype{EPFL-REPORT-200378}
\date{July}{22}{2014}
\newcommand{\comment}[1]{}
\newcommand{\KB}{KernelBoost\xspace}
\newcommand{\IKB}{Improved KernelBoost\xspace}
\newcommand{\IKBshort}{IKB\xspace}

\newcommand{\bbk}{{\bf k}}

\newcommand{\cN}{{\mathcal{N}}}

\newcommand{\IR}{{\mathbb{R}}}

\newcommand{\bbx}{{\bf x}}

\setcounter{page}{1}
\pagenumbering{arabic}
\pagestyle{plain}

\newpage
\begin{abstract}
  In this Technical Report we propose a set of improvements with
  respect to the KernelBoost classifier presented
  in~\cite{Becker13b}.
  We start with a scheme inspired by Auto-Context, but that
  is suitable in situations where the lack of large training sets
  poses a potential problem of overfitting.
  The aim is to capture the interactions between neighboring image
  pixels to better regularize the boundaries of segmented regions.
  As in Auto-Context~\cite{Tu09} the segmentation process is
  iterative and, at each iteration, the segmentation results for the
  previous iterations  are taken into account in conjunction with the
  image itself. However, unlike in~\cite{Tu09}, we organize our
  recursion so that the classifiers can progressively focus on
  difficult-to-classify locations.
  This lets us exploit the power of the decision-tree paradigm while
  avoiding over-fitting.

  In the context of this architecture, KernelBoost represents a
  powerful building block due to its ability to learn on the score
  maps coming from previous iterations.
  We first introduce two important mechanisms to empower the
  KernelBoost classifier, namely pooling and the clustering of
  positive samples based on the appearance of the corresponding
  ground-truth. These operations significantly contribute to increase
  the effectiveness of the system on biomedical images, where texture
  plays a major role in the recognition of the different image
  components.
  We then present some other techniques that can be easily integrated
  in the KernelBoost framework to further improve the accuracy of the
  final segmentation.

  We show extensive results on different medical image datasets,
  including some multi-label tasks, on which our method is shown to
  outperform state-of-the-art approaches.
  The resulting segmentations display high accuracy, neat contours, and
  reduced noise.
\end{abstract}




\section{Introduction}

Many recent papers~\cite{Tu09,Munoz10,Montillo11,Kontschieder13}  have shown the
importance of using  {\it context} when segmenting biomedical  images.  It helps
avoid  large  segmentation  errors  without  having to  rely  on  \emph{ad  hoc}
regularization priors~\cite{Rao10,Lucchi13a},  which are  commonly used  but can
only improve  results to  a limited  extent.  Using context  in this  manner can
therefore be  understood as  learning to capture  more complex  interactions to
better
regularize the segmented regions.

\begin{figure}[th]
  \centering
  \begin{tabular}{ccc}
    \includegraphics[width=0.36\linewidth]{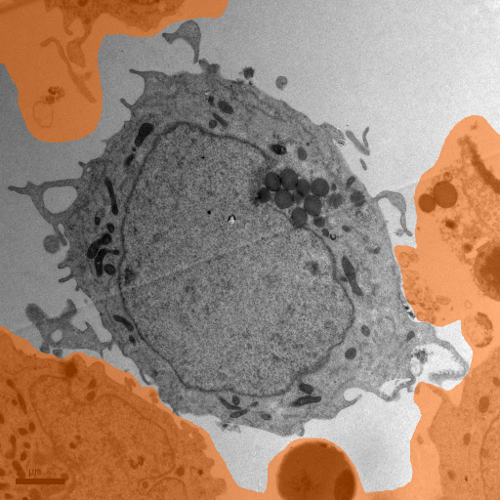} &
    \hspace{1em} &
    \includegraphics[width=0.36\linewidth]{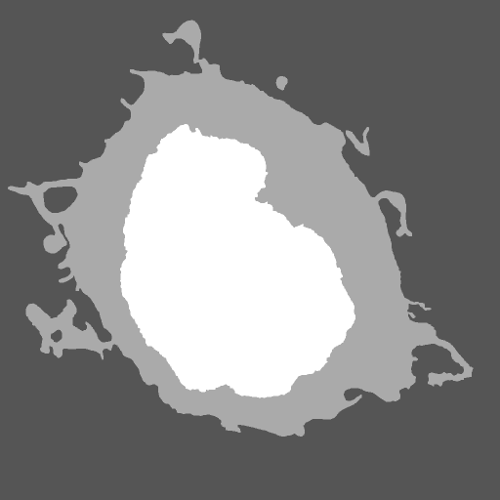} \\
    {\bf Original image} &
    \hspace{1em} &
    {\bf Ground-truth} \\
    \includegraphics[width=0.36\linewidth]{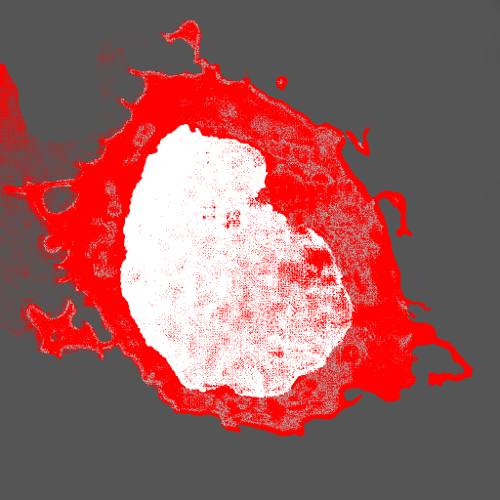} &
    \hspace{1em} &
    \includegraphics[width=0.36\linewidth]{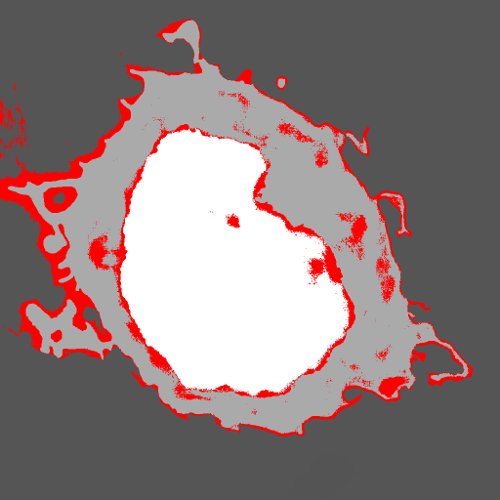} \\
    {\bf Random Forests} &
    \hspace{1em} &
    {\bf \IKB} \\
    \includegraphics[width=0.36\linewidth]{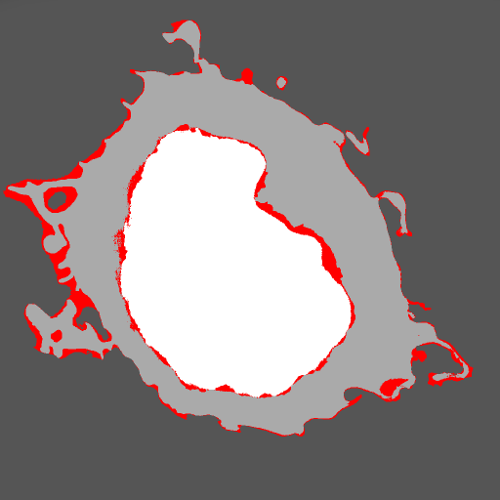} &
    \hspace{1em} &
    \includegraphics[width=0.36\linewidth]{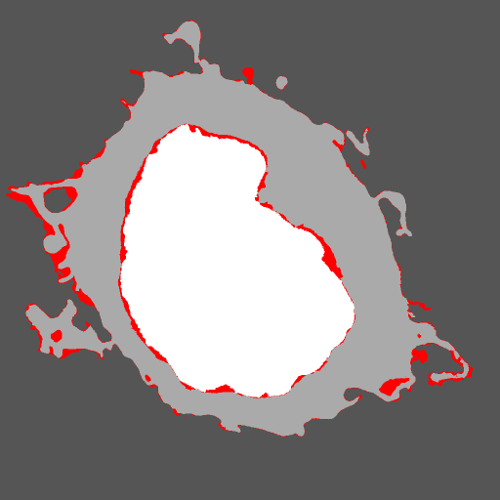} \\
    {\bf Auto-Context~\cite{Tu09}} &
    \hspace{1em} &
    {\bf Our approach} \\
  \end{tabular}
  \caption{Segmentation   example   for   a   test   image   from   the   Jurkat
    dataset~\cite{Morath13} (the area highlighted in orange is ignored during training/testing, as
    it includes cells that do  not appear in the  ground truth).  The thresholds  were selected to
    give, for each  method, the highest accuracy.  Note that  our approach finds
    accurate boundaries  between the different  regions, and that it  avoids the
    errors made by the other methods especially in the light gray region.
    Red overlays in the segmentations are used to mark the mistakes.
    Best viewed in color.}
  \label{fig:comparisons1}
\end{figure}

\begin{figure}[th]
  \begin{center}
    \includegraphics[width=0.98\linewidth]{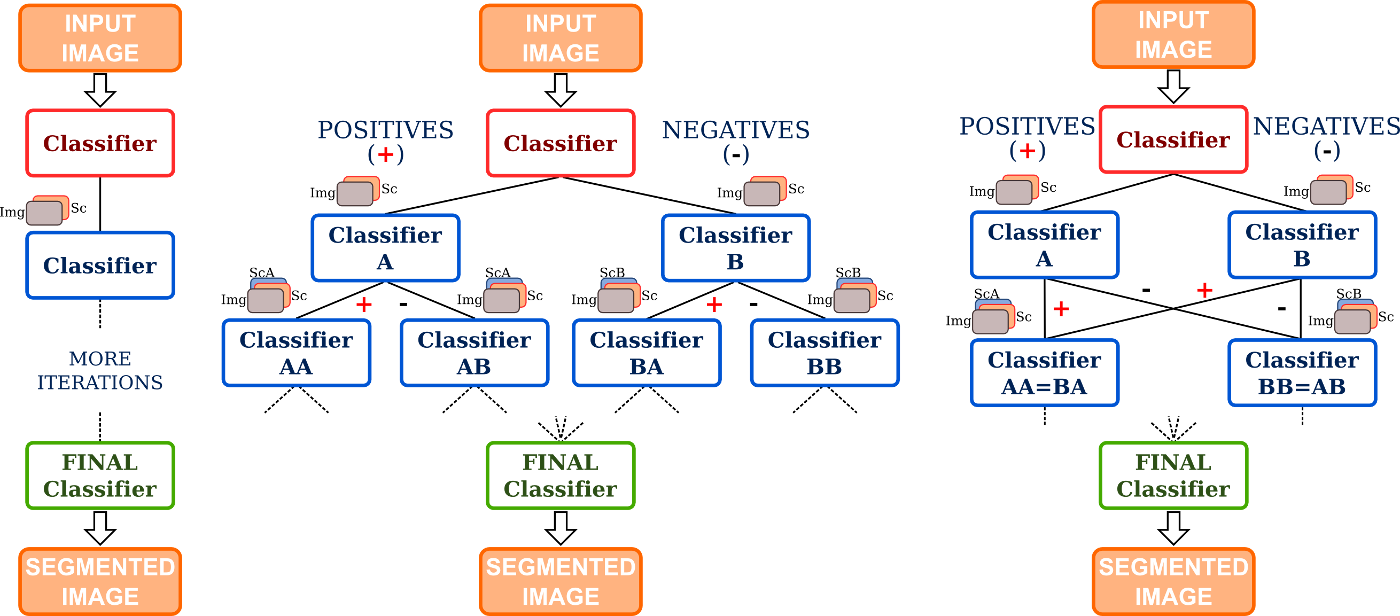}
  \end{center}
  \hspace{0.7cm} {\bf (a)} \hspace{3.4cm} {\bf (b)} \hspace{4.75cm} {\bf (c)}
  \caption{{\bf    (a)}   Schematic    representation   of    the   Auto-Context
    approach~\cite{Tu09}. At  each iteration, the segmentation  results from the
    previous iterations are  taken into account in addition to  the input image.
    {\bf  (b)}  Ideal  structure  of  our  system, which we have dubbed
    \emph{Expanded Trees}.
    The  output  of  the  first
    segmentation ({\it Sc}) is split  between positive and negative samples, and
    two separate  classifiers are  trained.  The procedure  is iterated  for the
    desired number  of levels, and  finally a classifier collects  the different
    output to  produce the  final segmentation. Since we  use powerful
    classifiers, many training samples are required, otherwise  we
    would  quickly run  out of  samples and  therefore overfit after
    few  iterations. Also, the number of classifiers to train -- and
    therefore the computational costs -- grows
    exponentially with the depth of the structure.
    {\bf (c)}  The approach  we propose.  We
    ``knot''  some
    branches together, and the classifiers in charge of a specific class at each
    level are  now the same.  This  avoids overfitting, as the
    classifiers are trained with larger training sets, while retaining  most of
    the classification  power of the tree  structure and keeping the
    computational effort limited.  Again, the output  of all
    the  intermediate steps  is  fed to  a classifier  that  produces the  final
    segmentation.  }
  \label{fig:scheme}
\end{figure}

In this  paper, we go one  step further by showing  that we can  use context not
only to avoid segmentation errors but also to accurately find region boundaries.
This  is challenging  for  several reasons.   First,  in most  kinds of  images,
classifiers trained to  differentiate locations belonging to one  kind of region
from those belonging  to another may struggle near  boundaries.  This is because
locations on both sides of the boundaries can be hard to distinguish, especially
when using feature  vectors computed using filters having  a substantial spatial
extent. Second, in biomedical images in general and particularly in those at the
scale of living cells, boundaries can have arbitrarily complex shapes.

To overcome these  difficulties we propose an approach that  is inspired by the
popular Auto-Context  algorithm~\cite{Tu09} but, unlike it,   progressively focuses  more and
more on \emph{difficult-to-classify} samples.

More specifically, for foreground/background segmentation purposes, Auto-Context
trains a chain of classifiers as depicted by Fig.~\ref{fig:scheme}(a): The input
to the first  classifier is simply the image data but  the next classifiers also
have  access to  the segmentation  results produced  by previous stages.  To
achieve our goal of focusing  on difficult-to-classify samples, we could instead
leverage the powerful decision  tree strategy~\cite{Breiman84}.  This would mean
recursively splitting  the pixels into those  that are classified  as foreground and
background  and  train  a  new  classifier  for  each  subset,  as  depicted  by
Fig.~\ref{fig:scheme}(b).  Unfortunately, unless the training set is
sufficiently large,
that would  be suboptimal because each  subsequent classifier would  have to be
trained on an ever smaller fraction of it, to the point where the training would
become ineffective.   This is particularly  true in the biomedical  field, where
often the amount of training data is limited because it is cumbersome to label.
Also, this would require the training of an exponentially growing
number of classifier, and therefore a considerable computational
effort. We will refer to this approach in the following as
\emph{Expanded Trees}.

As  shown  in  Fig.~\ref{fig:scheme}(c),  our  solution is  to  train  only  two
classifiers  at  each  stage.   The  first operates  on  the  samples  that  are
classified  as potential  foreground ones  by  {\it either}  classifiers at  the
previous stage, and the second one on samples classified as potential
negatives.
In this way, the  number of training examples used to  train the classifiers can
be kept roughly constant at all stages, which avoids overfitting.
Since we ``knot'' the two branches of the tree, we have dubbed this
approach \emph{Knotted Trees}.
To focus  on  the difficult-to-classify  examples,  we use  permissive
bounds to decide if a sample is  potentially foreground or background. For
example, a sample  classified as negative but  with a low confidence  score is a
difficult-to-classify sample, and it is sent  to the two classifiers at the next
stage. In this way,  both classifiers  will have  access  to these  important
samples.  A final  classifier collects  the partial  results from  the different
branches and outputs the final segmentation.

In the evaluation of the schemes we presented, we use in each node a KernelBoost
classifier~\cite{Becker13b}. However, other classifiers
can be used as long as they are powerful enough, at least in the earlier stages.

Our approach naturally extends  to multi-label segmentation  problems.  We
simply train our approach for each  label in a 1-versus-1 scheme, where separate
binary problem are  set up for each  pair of classes. Again,  a  final
classifier collects the partial results  from the different branches and outputs
the final segmentation.

We use  four different  datasets to demonstrate  that our  approach outperforms
both Auto-Context~\cite{Tu09} and a recent CRF-based method~\cite{Lucchi13a}, and
to assess the strengths and weaknesses of the methods we propose.


\section{Related Work}

Image segmentation algorithms  typically rely on  local image cues combined with spatial
constraints.    For  example,   Markov  Random   Fields~(MRFs)  use
 unary  terms that depend  on image features together  with pairwise
potentials  that  enforce  simple smoothness  priors~\cite{Li95,Perez98}.   With
Conditional  Random  Fields~(CRFs),  the  smoothness  terms  can  also  be  made
dependent on the image cues~\cite{Sutton07,Plath09}.   This can exploit the fact
that  boundaries between  image  regions have  specific  image appearance,  and
significantly improves the quality of the resulting segmentation.

However  it is  not  entirely clear  that simple  spatial  features truly make a difference
 when powerful image features are used, and several authors report good
results         even     without        spatial
constraints~\cite{Ramanan06b,Verbeek07,Shotton09,Lucchi11a}.       Higher     orders
terms~\cite{Kohli08} or hierarchical  approaches~\cite{Ladicky09} have therefore
been introduced  to capture more global  constraints.  Unfortunately, optimizing
the  parameters of  such complex  models  in CRFs  quickly becomes  intractable.
\cite{Nowozin13a} shows that decision trees can  be used in Decision Tree Fields
to  model both  the unary  and  the pairwise  terms.  This  makes the  parameter
estimation problem  much more tractable.  \cite{Lucchi13a}  relies on subgradient
descent to efficiently learn CRF models  for segmentation using a working set of
constraints   in   a   Structured   SVM  formulation.    This   method   obtains
state-of-the-art  results on  medical data,  and we  compare against  it in  the
Experimental Results section.

However despite the widespread interest  in them, CRFs are still computationally
expensive  at  run-time.   A  more   efficient  approach  to  enforcing  spatial
constraints was  introduced in~\cite{Tu09} under  the name of  Auto-Context.  In
Auto-Context,   a  first   segmentation   is  obtained   by  simple   pixel-wise
segmentation, followed by a second  segmentation   obtained  by combining  image
features with ``context features''. They are similar to
image features  but are computed on  the output of  the first iteration.
This process can be iterated  until convergence.
This scheme was inspired by~\cite{Loog06}.
  A similar
approach was also developed in~\cite{Shotton09} together with a CRF model.    Entangled Forests~\cite{Montillo11}
use similar features  in the nodes of  Random Forests, applied to  the output of
previous nodes.  Geodesic Forests~\cite{Kontschieder13} extend these features to
depend on geodesic distances between pairs of locations to exploit long-distance
correlations.  However  using the geodesic  distances assumes that  strong image
gradients are highly correlated with boundaries between regions, which is often not
true in medical images, as shown in Figs.~\ref{fig:comparisons1},~\ref{fig:texture}, and~\ref{fig:data_dna}.

We also use features computed from previous segmentations. However we show how to
leverage this general approach to significantly improve the quality of the final
segmentation, by focusing on the difficult-to-classify locations.

The idea of splitting samples according to the estimated probabilities
was already proposed in~\cite{Tu05}, but they adopt a traditional
binary tree structure, they do not pass the estimated probabilities
down the tree to enrich the input features -- therefore they lack context
information --, they do not learn features on the newly estimated
probabilities, and they do not have a final classification stage.


\section{Our Approach}

In  this section  we  describe  our approach. It relies on a  statistical
segmentation method that can take as input the original image and the results of
the  previous  segmentation   steps.   In  practice  we   use  the  KernelBoost
method~\cite{Becker13b}, which we improved for this work. We therefore
describe it in Section~\ref{sec:KB_and_IKB} for completeness, along with our modifications.

\subsection{Knotting Branches for Segmentation}

In our  approach, as illustrated in Fig.~\ref{fig:scheme}(c), we  first segment the input image  using a
classifier that  predicts for each  image location  to which
class  the location  belongs,  together  with a  confidence  measure about  this
prediction.

In  general this  first  segmentation has margins for improvement, and  we perform
subsequent segmentations that rely not
only on the original image but also on the output of previous steps.

This is similar to what Auto-Context does, but differs in a critical way: we
separate image locations into a ``positive set'' made of  the ones classified as
positives, and a ``negative set'' made of the ones classified as negatives. Each
set is then processed independently by a new classifier. We iterate this process
until convergence on the training set.   The output of the different classifiers
are then fed to a final classifier to produce the final segmentation.

\newcommand{\trainingset}{\mathcal{T}}
\newcommand{\image}{{\bf X}}
\newcommand{\labels}{{\bf Y}}
\newcommand{\classifier}{\varphi}
\newcommand{\wh}[1]{\widehat{#1}}
\newcommand{\wb}[1]{\overline{#1}}
\newcommand{\loss}{L}
\newcommand{\Pos}{P}
\newcommand{\Neg}{N}

More  formally, let  $\trainingset  = \{(\image_i,  \labels_i)\}_i$ be a set  of
training images $\image_i$  together with the corresponding  ground truth labels
$\labels_i$. We write
\begin{equation}
\labels_i(u,v) = \left\{
\begin{array}{cc}
+1 &  \text{ if } (u,v) \text{ in }\image_i\text{ belongs to foreground,}\\
-1 & \text{ otherwise.}\\
\end{array}
\right.
\end{equation}
We train a first classifier
\begin{equation}
\classifier^0    =   \argmin_\classifier    \sum_i   \sum_{(u,v)}
\loss\big(\labels_i(u,v),\classifier(\image_i)(u,v)\big) \> ,
\end{equation}
where $L$ is a loss function.

$\classifier^0(\image)$   can   be   seen   as   a   confidence   map:
$|\classifier^0(\image)(u,v)|$  is a  confidence measure  for location
$(u,v)$ in $\image$.
However, to avoid locations  with large absolute
values bias the further steps, we  use instead a version normalized to
lie in the $[-1,+1]$ by considering
\begin{equation}
  \wb{\classifier}^0(\image) = 2p\left( \labels_i = 1
  \left|\right.\image_i\right) -1,
\label{eq:normalisation}
\end{equation}
where $p\left( \labels_i = 1 \left|\right.\image_i\right)$ is the probability of a pixel of being in
the positive class. We use
\begin{equation}
  p\left( \labels_i = 1
  \left|\right.\image_i\right) = \frac{1}{1+\exp(-2\classifier^0(\image))}
\label{eq:prob_norm}
\end{equation}
as in~\cite{Hastie01}.

We then define the positive $\Pos_i^0$ and negative $\Neg_i^0$ sets for each
training image $\image_i$ as
\begin{equation}
\begin{array}{ll}
\Pos_i^0 & = \left\{ (u,v) \> | \> \wb{\classifier}^0(\image_i)(u,v) > -\varepsilon
\right\} \> , \text{ and}\\
\Neg_i^0 & = \left\{ (u,v) \> | \> \wb{\classifier}^0(\image_i)(u,v) < +\varepsilon
\right\} \> .\\
\end{array}
\end{equation}
Because  we use  $\varepsilon  >  0$, the  positive  sets $\{\Pos_i^0\}_i$  contain
locations  that are  classified  by $\wb{\classifier}^0$  as  positive but  also
locations classified as negative but with low confidence. In practice,
we use $\varepsilon = \frac{1}{2}$. Similarly, the negative sets $\{ \Neg_i^0 \}_i$ contain
locations classified  as positive with low confidence.  Both the
positive  and  negative sets  therefore  contain  all the  difficult-to-classify
samples, which will be treated by two new classifiers $\classifier_\Pos^1$ and
$\classifier_\Neg^1$. The first is taken to be
\begin{equation}
\begin{array}{ll}
\classifier_\Pos^1    = &  \argmin \limits_\classifier    \sum\limits_i   \sum\limits_{(u,v)\in\Pos_i^0}
\loss\big(\labels_i(u,v),\classifier(\image_i^1)(u,v)\big) \> ,\\
\end{array}
\end{equation}
where     $\image_i^1      =     \big(\image_i,\wb{\classifier}^0(\image_i)\big)$.
$\classifier_\Pos^1$ is therefore trained on  the locations in the positive sets
only,    and     has    access    to    the     normalized    confidence    maps
$\wb{\classifier}^0(\image_i)$ in addition to the  training images.  Its task is
thus simpler than that of the first classifier.  The $\classifier_\Neg^1$ classifier
is defined in a similar way.

New positive and negative sets,  respectively noted $\Pos_i^1$ and $\Neg_i^1$, are
computed as before  from the output of the  two classifiers $\classifier_\Pos^1$
and  $\classifier_\Neg^1$.
Next,  we  train   a  second  pair  of  classifiers
$\classifier_\Pos^2$   and   $\classifier_\Neg^2$.    The   $\classifier_\Pos^2$
classifier is taken as:
\begin{equation}
\begin{array}{ll}
\classifier_\Pos^2 = & \argmin \limits_\classifier \sum\limits_i \sum\limits_{(u,v)\in\Pos_i^1} \loss\big(\labels_i(u,v),\classifier(\image_i^{2,P})(u,v)\big) \> ,\\
\end{array}
\end{equation}
where $\image_i^{2,P}$ is made of image $\image_i$, and all the corresponding confidence maps
computed up to this point: $\image_i^{2,P} =
\big(\image_i,\wb{\classifier}^0(\image_i), \wb{\classifier}_\Pos^1(\image_i^1) \big)$.
Similarly,
\begin{equation}
\begin{array}{ll}
\classifier_\Neg^2 = & \argmin \limits_\classifier \sum\limits_i \sum\limits_{(u,v)\in\Neg_i^1} \loss\big(\labels_i(u,v),\classifier(\image_i^{2,N})(u,v)\big) \> ,\\
\end{array}
\end{equation}
This process is iterated as long as the numbers of misclassified samples in both
branches of  our ``tree'' remain above  a threshold.

Finally we  train a Randomized Forest  to predict the locations  labels based on
all   the    available   data, including the input image, feeding the classifier with a feature descriptor
\begin{equation}
\Delta = \Big\{\Big(\big(\image_i,   \wb{\classifier}^0(\image_i),
\wb{\classifier}_\Pos^1(\image_i^1),        \wb{\classifier}_\Neg^1(\image_i^1), \cdots
\big), \labels_i\Big)\Big\}_i.
\end{equation}
To incorporate additional information about a pixel's neighborhood, we
also consider the values located on the corners and on the middle of
the side of two nested squares centred on the point, with side of 5
and 10 pixels respectively, in a stylized snowflake configuration.

In the case of a multi-label segmentation problem, we proceed as described above
for  each label  in a  1-versus-all scheme.   Again, a  final Randomized  Forest
is trained to predict the label based on all the intermediate data created for
all the labels.

At run-time, we simply have to apply to the input image $\image$ the successive
classifiers $\classifier^0$, $\classifier_\Pos^0$, $\classifier_\Neg^1$,
$\cdots$ to compute the normalized confidence maps, which are then fed to the final
Randomized Forest to obtain a final confidence map.

\subsection{Expanded Trees}
In the case of the ideal structure of Fig.~\ref{fig:scheme}(b), we
proceed as explained above for what concerns the splitting criteria,
but instead of sending samples to the parallel branch when they meet
the criteria to switch branch, we simply continue the splitting
process.
When the number of samples in a node falls below a critical threshold,
we stop the growing of that branch.

\section{KernelBoost, Improved KernelBoost, and extensions}
\label{sec:KB_and_IKB}

KernelBoost (KB) is a  statistical method that segments an image  by classifying each
pixel independently  into two classes.
It has the key advantage of not requiring to hand-design the kernels
it operates with, and that of having very few, easily tunable parameters.

The  training data is a  set of training
samples $\{(\bbx_i,y_i)\}_{i=1  ...  N}$, where  $\bbx_i \in \IR^n$ is  an image
patch and $y_i  \in \{-1,1\}$ its corresponding label. From  this set, KernelBoost first
generates  a  bank   of  discriminative  kernels,  and  then   builds  a  binary
GradientBoost classifier,  with weak learners  of the form of  regression trees.
Each node  of these  trees contains a  test selected from  the kernel  bank.  At
run-time,  the selected  kernels $\bbk_j$  can simply  be used  as convolutional
filters applied to the image to segment, which makes the approach efficient.

\subsection{Pooling and training sample clustering}

\begin{figure}[t]
  \centering
  \begin{tabular}{ccc}
    \includegraphics[width=0.3\columnwidth]{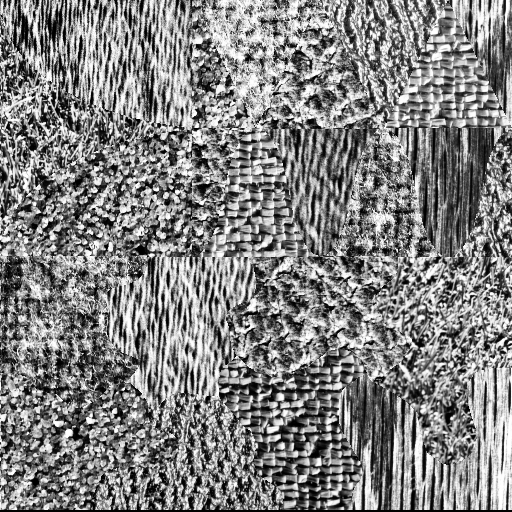} &
    \hspace{1em} &
    \includegraphics[width=0.3\columnwidth]{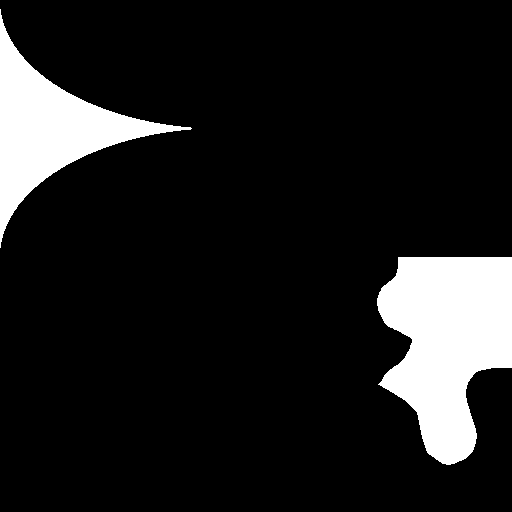} \\
    {\bf (a)} &
    \hspace{1em} &
    {\bf (b)} \\
    \includegraphics[width=0.3\columnwidth]{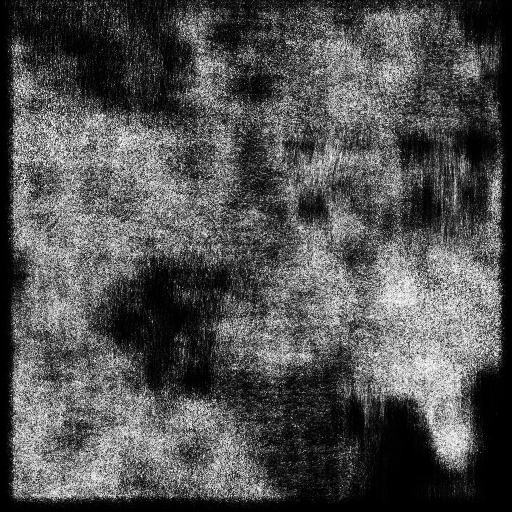} &
    \hspace{1em} &
    \includegraphics[width=0.3\columnwidth]{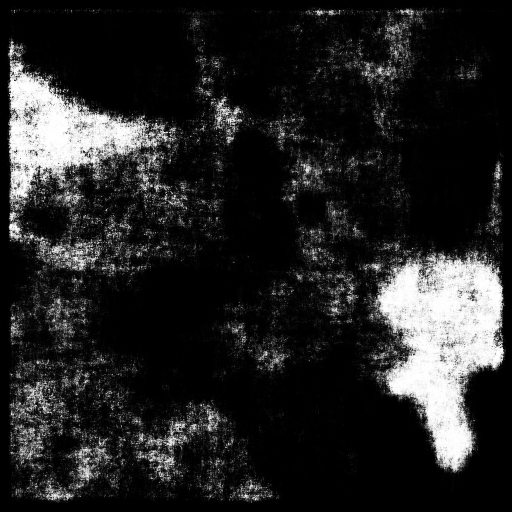} \\
    {\bf (c)} &
    \hspace{1em} &
    {\bf (d)} \\
  \end{tabular}
  \caption{Texture  experiment.  The segmented  image  is  a mosaic  of  Brodatz
    images,  created at  USC-SIPI  for research  on  texture segmentation.   The
    dataset             is              available             at             the
    address~\url{http://sipi.usc.edu/database/?volume=textures}.        {\bf(a)}
    Original  test pattern.   {\bf(b)} Ground-truth  for the  selected material.
    {\bf(c)}  Segmentation  obtained   by  the  KernelBoost  algorithm~\cite{Becker13b}.
    {\bf(d)}  Segmentation  obtained  by  the KernelBoost  algorithm  coupled  with  the
    POSNEG/MAX-pooling scheme.}
  \label{fig:texture}
\end{figure}

While  obtaining  excellent  performances  in  the  delineation  of  curvilinear
structures~\cite{Becker13b}, the original KernelBoost method lacks a mechanism
for dealing with  more complex patterns, such as the  ones that typically appear
in medical images.   To highlight this deficiency,  we applied the original  KernelBoost to a
texture  segmentation  problem.   As  shown  in  Fig.~\ref{fig:texture}(c),  the
quality of the segmentation is quite poor.

This is because, for texture recognition  problems, the filter responses
should be made robust  to slight shifts in the images.   We therefore introduce a
pooling  mechanism~\cite{Boureau10b} over  the  filter  responses. We  evaluated
several possible mechanisms, and we empirically noticed that keeping the maximum
value    of   the    responses   of    the POSNEG operator~\cite{Rigamonti11b}
over a region performed best.

The  POSNEG operator  transforms  the result  of a  convolution  $C$ into  two
images,  one made  of the  positive values  of $C$,  the other  one made  of the
opposite of the  negative values of $C$.  The optimization  over each node finds
which filter and which of these two images should be used for best performances.

Moreover,  the kernels  learned  by  the original KernelBoost filters  are not  always
optimal. This  is because they  are learned in  a discriminative fashion  in the
attempt to  distinguish all the  positive samples  from the negative  ones.
However,
this  leads to  suboptimal performances  because  positive samples  can be  very
different from each other, for example in presence of texture.

We therefore cluster  the positive samples based on their  appearances, and each
kernel is generated by classifying the  positive samples of one cluster randomly
chosen  against all  the negative  samples.  We  empirically observed  that this
second  modification also  improves  the  results  compared to  the
original KernelBoost.

However,  even with  these improvements, KernelBoost can  make significant mistakes if
not incorporated in the framework we propose, as
shown in Fig.~\ref{fig:comparisons1}.
In the following we will refer to KernelBoost, when both pooling and clustering are used,
as \emph{\IKB} (\IKBshort for short).

\subsection{Superpixel pooling and superpixel features}
Superpixels~\cite{Felzenszwalb04} emerged as an efficient way of
dealing with large images and image stacks, while at the same time
improving the performances by averaging the results of pixel-based
approaches over small, homogeneous image regions.
The idea behind them is to group meaningful regions and to use
them as a substitute of the regular grid structures imposed by
pixels~\cite{Achanta12}.
The same pixel-based regular grid is used by traditional pooling
operators available in literature, we have therefore considered
replacing it with a structure based on superpixels.
However, given the homogeneity of the pixels captured inside each
superpixel, pooling the feature values inside each superpixel would be
of little use.
Indeed, the main goal of the pooling step is to introduce robustness
against small distortions, translations and rotations, and this
requires the pooling scheme to be independent from the underlying
data.

\begin{figure}[t]
  \centering
  \begin{tabular}{cc}
    \includegraphics[width=0.4\columnwidth]{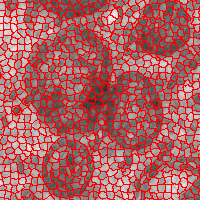} &
    \includegraphics[width=0.4\columnwidth]{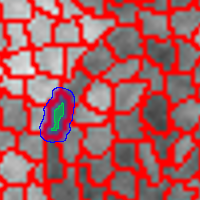} \\
    {\bf (a)} &
    {\bf (b)} \\
  \end{tabular}
  \caption{Analysis of the regions identified by SLIC
    superpixels~\cite{Achanta12} on a fragment of a medical image.
    {\bf(a)} SLIC
    superpixels extracted by the code provided
    by~\cite{Achanta12}. Notice that they accurately distinguish
    different regions, such as the membranes of the mitochondria.
    {\bf(a)} Example extended/shrinked regions --- in blue and green
    respectively --- where pooling can be performed to extract
    meaningful information about the mitochondrion's membrane.}
  \label{fig:mito_sp}
\end{figure}

The idea of analyzing filter responses over structures whose shapes
are not imposed by rigid rules, but by the image itself, is nonetheless
appealing.
We have therefore considered not the superpixels extracted from the
images, but the regions
which can be computed from them by erosion and dilation by a
given amount of pixels.
If we consider, for instance, the superpixel segmentation
of Fig.~\ref{fig:mito_sp}(a), we can observe that it is able to group
interesting regions, such as the membranes of the mitochondria.
While pooling directly on them would simply average very similar
filter responses on an homogeneous region, if we consider smaller and
larger regions we can hope that, for instance, the classifier gets fed
with the border of each membrane, easing its recognition.
As an example, this can happen by considering the two regions,
outlined in green and blue respectively, of
Fig.~\ref{fig:mito_sp}(b).
We have dubbed this approach \emph{superpixel pooling}.

In the KernelBoost context we can go even further and perform
operations over different regions computed from the individual
superpixels.
The classifier present in each weak learner will be fed with a
set of features resulting from potentially nonlinear operations over
multiple regions --- for instance the difference of the
absolute values of the negative parts of the two regions in
Fig.~\ref{fig:mito_sp}(b) --- and will then be able to select the most
discriminative combinations for a given training set.

Among the different approaches available in literature, SLIC
superpixels~\cite{Achanta12} emerged for their simplicity and their
capability to adhere to image boundaries.
We have therefore based our
proposal on the superpixels extracted by this approach.

\subsection{External features}
KernelBoost is particularly well suited to exploit the power of
external hand-crafted or learned features.
It can indeed load them as additional channels, and perform filter
learning on them, thus empowering them by building a two-level
architecture.
A particularly interesting example of features is constituted by
Ilastik features~\cite{Sommer10}. They are constituted by multiple
hand-crafted features --- such as Gaussians, Structure Tensor, \ldots
--- computed with different parameter settings, for instance different
Gaussian smoothing.
They are used in many state-of-the-art segmentation tools used by
practitioners.

The features that can be used are usually dataset-dependent.
For instance, in one dataset where the task is the segmentation of
mitochondria, we have included as an additional channel the
mitochondria membranes obtained by the algorithm of~\cite{Sironi14},
and this proved to significantly improve the final performance, as
shown in Section~\ref{sec:ress}.

\subsection{Multiple scales}
When dealing with medical images, the structures of interest commonly
appear at very different scales or occupy relevant portions of the
images.
In typical EM images of mitochondria, for example, a single
mitochondrion might be  several hundreds pixels in length.
For this reason, learning filters from data at full image scale
might lead to a set of filters which are able to capture only small
fractions of the objects, engendering errors in a
pixel-level segmentation.

The KernelBoost approach can be easily extended to deal with multiple
scales, given its ability to operate on separate input channels.
We just have to incorporate the notion of scale in the sampling scheme,
to ensure that the same sample is collected over the different scales
considered.
The filter learning process is then performed independently for each
individual channel at the proper scale, and the regression tree
learning scheme will then be in charge of selecting which filter and
which scale are more discriminative.

\subsection{Fake-3D}
EM imagery often produces complete, anisotropic image stacks.
Full 3D information can be very useful in discriminating the different
components of the scene, but it is usually computationally expensive
and memory intensive to deal with.
Moreover, the anisotropicity of the data might result in degenerate
filters and therefore poor performance.
For these reasons, while KernelBoost could in principle operate on N-D
data --- experiments with 3D image stacks are presented
in~\cite{Becker13b} --- we have chosen to operate on 2D data but
incorporating data from multiple neighboring slices.
In particular, the whole segmentation scheme operates on individual
slices until it reaches the final classifier stage.
At that point the classifier is fed not only with the features
extracted on the considered slice, but also with the features
extracted on a slice $D$ steps before and on a slice $D$ steps after
it.
Padding is performed to deal with stack's boundaries.
We have considered in our experiments $D=3$, but its exact value has
to be tuned according to the size of the structures of interest and
the inter-slice distance.

\subsection{Z-cut}
Biomedical image stacks usually contain components with no predominant
orientation. The choice of the axes is therefore arbitary.
When enough slices are available, it might be interesting to consider
not only the images in the traditional X-Y plane, but also in the X-Z
and Y-Z ones.
This is particularly useful in the context of an architecture such as
those presented in the previous chapter, where different levels of
recursion can focus on different image planes.
Indeed, if the anisotropicity is not too elevated, filters can impose
a regularity on the structures found in these planes, giving a final
3D result which is smoother and less noisy.


\section{Experimental Results}
\label{sec:ress}
To validate our  approach we  performed  extensive experiments on
four very different medical image datasets.

\subsection{The Human T-Cell Line Jurkat Dataset~\cite{Morath13}}
\label{sec:t_cell}

The  first dataset  we consider is composed by Transmission Electron
Microscopy (TEM) images of
the  human T-cell  line Jurkat~\cite{Morath13}.   We   randomly selected  10
training images and 4 test images, each with size $1024\times 1024$
pixels.  We  then created  a set  of masks to  ignore, both  in training  and in
testing, the  components that are  not considered in the  available ground-truth
images.  The particularity  of this dataset is that it  contains three different
classes, for the  background, the cytoplasm, and the nuclei.   A sample training
image,   along   with   the   corresponding   ground-truth,   is   depicted   by
Fig.~\ref{fig:comparisons1}.
The goal of the final user of this data is to estimate, with the higher accuracy
possible, the areas of the different components of the cells, in order
to use these values to tune numerical models of the response of
illnesses to different treatments.

The main challenge of this dataset is the fact that the cytoplasm and the cell's
nucleus     look    very     similar.
The     results  obtained by the different approaches  are     given    in Table~\ref{tab:jurkat}.
Our proposal  significantly outperforms the other
methods, including that  of~\cite{Lucchi11b}.
Visual    inspection    of     our    approach,    as    seen    in
Fig.~\ref{fig:comparisons1}, shows that  not only the accuracy  score is higher,
but  also  the contours  are  significantly  better delineated  and  macroscopic
mistakes get fixed.

\begin{table}[t]
  \caption{ Results  for the  Jurkat dataset~\cite{Morath13}.   The segmentation
    accuracy  is computed  as  the  fraction of  pixels  whose  label match  the
    ground-truth data. See Section~\ref{sec:t_cell} for more details.}
  \label{tab:jurkat}
  \centering
  \begin{tabular}{@{}lc@{}}
    \toprule[0.1em]
    \textbf{Method} &  {\bf Accuracy}\\
    \midrule
    Random Forests                                & 0.753 \\
    Lucchi~\etal~\cite{Lucchi11b}                 & 0.836 \\
    \IKB                              & 0.929 \\
    Auto-Context~\cite{Tu09} (on \IKB)     & 0.956 \\
    {\it our approach} (on \IKB), no norm. & 0.941 \\
    {\it our approach} (on \IKB)           & {\bf 0.973} \\
    \bottomrule[0.1em]
  \end{tabular}
\end{table}


Table~\ref{tab:jurkat} also shows that normalization of the classifiers' output as expressed
by Eq.~\eqref{eq:normalisation}  is an important  step, as the results  without it
are not significantly better than the ones obtained with the original \IKB method
on  this dataset.

\begin{figure*}[ht]
  \centering
  \begin{tabular}{ccc}
    \includegraphics[width=0.32\linewidth]{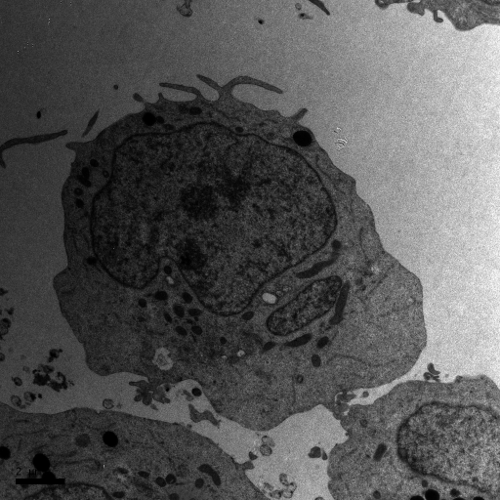} &
    \includegraphics[width=0.32\linewidth]{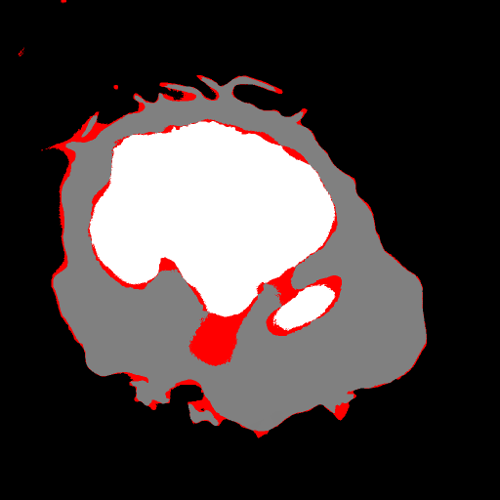} &
    \includegraphics[width=0.32\linewidth]{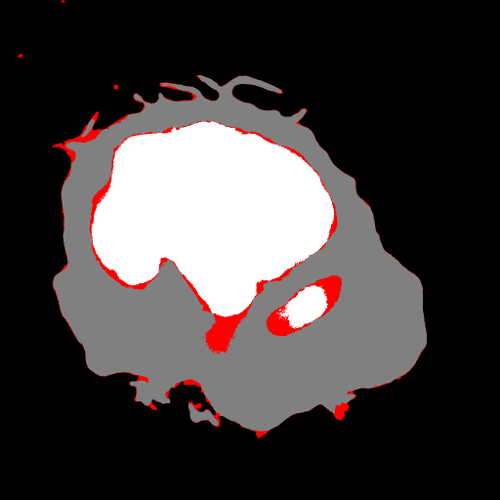} \\
    {\bf Original image}
    & {\bf Auto-Context~\cite{Tu09}} & {\bf Our approach} \\
  \end{tabular}
  \caption{Segmentation errors for a  test   image   from   the   Jurkat
    dataset~\cite{Morath13}. The errors are highlighted in red in the images.
    Our approach significantly reduces the mistakes --- the accuracy for our approach is 0.972, while it is 0.956 for Auto-Context.
    Best viewed in color.}
  \label{fig:jurkat_errors}
\end{figure*}

Finally, we have applied the Auto-Context approach~\cite{Tu09} on the
output of the \IKB method.
While Auto-Context improves the segmentation results for this dataset,
the improvement is smaller than that achieved by our strategy.
Fig.~\ref{fig:jurkat_errors} shows how errors are distributed for
different approaches on a test image.
Note also that what we have labeled \textit{Auto-Context} is an improved
version of the algorithm of~\cite{Tu09}, since filters are learned at
each stage on the newly added score images, and the final set of score
maps is used to feed a Random Forest classifier instead of simply
relying on the last score map.
In our tests with methods that do not learn filters on the feature
maps, for instance~\cite{Becker12}, the improvements linked with the
use of Auto-Context are negligible.

To validate our results, we have randomly selected another set of 10
training and 6 testing images --- denoted here as \emph{Jurkat-B
  dataset} ---, and we have reported the results
obtained by the three methods of Fig.~\ref{fig:scheme} in
Table~\ref{tab:jurkatB}.
As it can be seen, our approach performs better then Auto-Context ---
which is reasonable and confirms the results obtained in the other
random split --- but also the Expanded Trees, since the amount of
training images is limited and therefore the system might suffer
because of that.

\begin{table}[t]
  \caption{
    Results for the Jurkat-B dataset~\cite{Morath13}.
  }
  \label{tab:jurkatB}
  \centering
  \begin{tabular}{@{}lc@{}}
    \toprule[0.1em]
    \textbf{Method} & {\bf Accuracy} \\
    \midrule
    KernelBoost                       & 0.909 \\
    Auto-Context~\cite{Tu09} (on KernelBoost)                   & 0.922 \\
    Expanded Trees (on KernelBoost)                            & 0.934 \\
    {\it Knotted Trees} (on KernelBoost)                       & {\bf 0.948} \\
    \bottomrule[0.1em]
  \end{tabular}
\end{table}



\subsection{NIH-3T3 Fibroblast Cells Dataset~\cite{Coelho09}}
\label{sec:fibroblast}

\begin{figure}[t]
  \centering
  \begin{tabular}{ccc}
    \includegraphics[width=0.38\columnwidth]{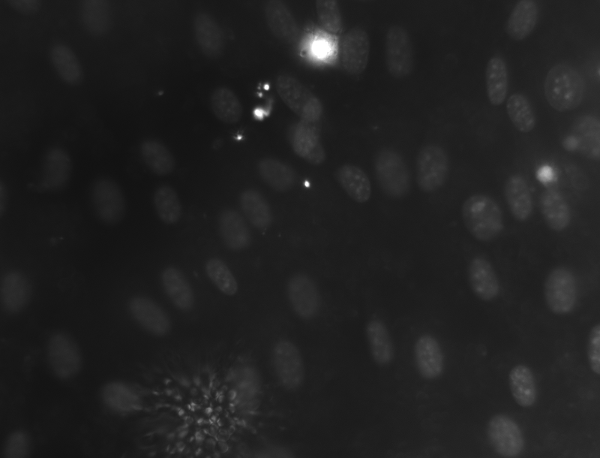} &
    \hspace{1em} &
    \includegraphics[width=0.38\columnwidth]{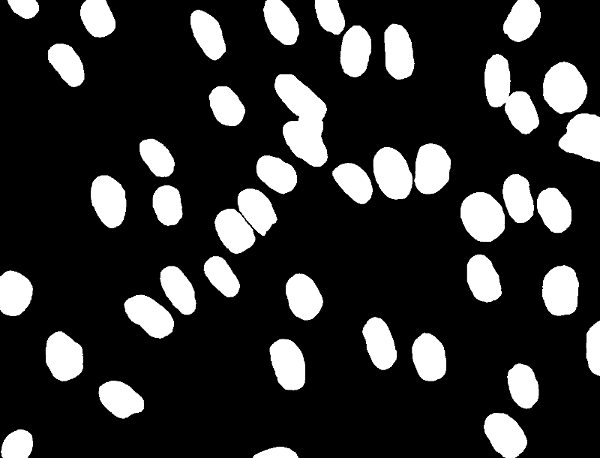}  \\
    {\bf (a)} &
    \hspace{1em} &
    {\bf (b)} \\
    \includegraphics[width=0.38\columnwidth]{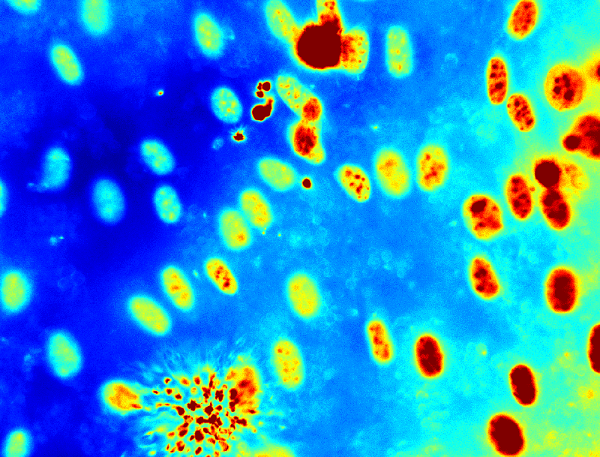} &
    \hspace{1em} &
    \includegraphics[width=0.38\columnwidth]{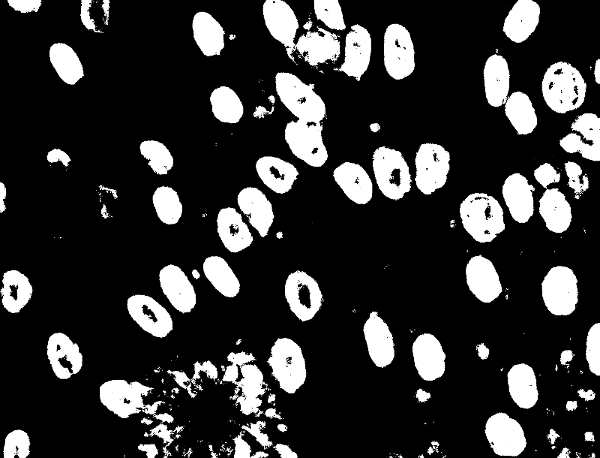} \\
    {\bf (c)} &
    \hspace{1em} &
    {\bf (d)} \\
    \includegraphics[width=0.38\columnwidth]{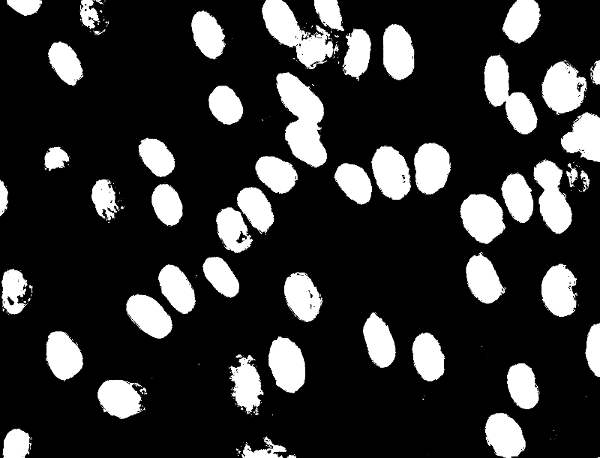} &
    \hspace{1em} &
    \includegraphics[width=0.38\columnwidth]{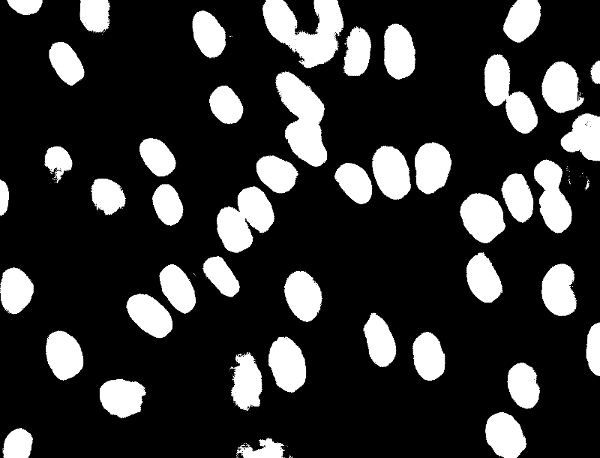} \\
    {\bf (e)} &
    \hspace{1em} &
    {\bf (f)} \\
  \end{tabular}
  \caption{Segmentation  results for  a  randomly-selected test  image from  the
    NIH-3T3 fibroblast  cells dataset~\cite{Coelho09}.   The thresholds  used to
    obtain the segmentations from the score  images are given by the values that
    maximize the VOC  score for the given image.  {\bf(a)}  Original test image.
    The poor contrast exhibited by the images is one of the characteristics that
    make this dataset difficult.  {\bf(b)}  Ground-truth for the selected image.
    {\bf(c)} Original  test image, after  being manually altered to  improve the
    visual quality.  Best viewed in colors.  {\bf(d)} Segmentation obtained with
    \IKB.  The  VOC score is 0.7243.  {\bf(e)} Segmentation  obtained by our
    approach except that the samples are split into two balanced sets instead of
    a  positive  and  negative  sets.    The  VOC  score  is  0.8024.   {\bf(f)}
    Segmentation obtained by our approach. The VOC score is 0.8160.  }
  \label{fig:data_dna}
\end{figure}

The second  dataset we consider  is  constituted by  2D fluorescence  microscopy
images   of   Hoechst   33342-stained   NIH-3T3   mouse   embryonic   fibroblast
cells~\cite{Coelho09}.  As it  can be seen in the sample  test image depicted by
Fig.~\ref{fig:data_dna}, the  images often  contain debris  and the  nuclei vary
greatly in brightness, making their segmentation challenging.  For these reasons,
this  dataset was  chosen for  one  of the  competitions  in the  ISBI2009
challenge.  The training  set is composed by 4 randomly  chosen test images with
size $1344\times  1024$ pixels,  while the  test set  contains the  remaining 45
images.

Table~\ref{tab:DNA} shows that our approach,  using our \IKB for the
$\classifier$  classifiers,   significantly  outperforms   the  state-of-the-art
algorithm  for this  dataset~\cite{Song13},  as  well as  the  other methods  we
evaluated. Our  approach improves on the starting point, given by \IKB,  which  in turn is  better  than  the
original \KB classifier. It is also better
than Auto-Context, which does not bring a significative improvement
over the starting point.


\begin{table}[t]
  \caption{ Segmentation  accuracy for the NIH-3T3  fibroblast cells
    dataset~\cite{Coelho09}.  The results for  the human expert were
    obtained by a second human expert on a subset of 5 images of the
    test set,  and are given  only to demonstrate the  complexity of
    the task. See Section~\ref{sec:fibroblast} for more details.}
  \label{tab:DNA}
  \centering
  \begin{tabular}{@{}lccc@{}}
    \toprule[0.1em]
    \textbf{Method} & {\bf VOC} & {\bf RI}  & {\bf DI} \\
    \midrule
    Human expert~\cite{Coelho09}         & -            & 0.93        & - \\
    Song~\emph{et al.}~\cite{Song13}     & 0.852        & 0.932       & 0.906 \\
    \KB                 & 0.817        & 0.921       & 0.899 \\
    \IKB                                 & 0.833        & 0.929       & 0.909 \\
    Auto-Context (on \IKB)        & 0.842        & 0.920       & 0.914 \\
    {\it our approach (on \IKB)}  & {\bf 0.874}  & {\bf 0.946} & {\bf 0.932} \\
    same, but with 50\%-50\% split       & 0.842        & 0.906       & 0.914 \\
    \bottomrule[0.1em]
  \end{tabular}
\end{table}


We also investigated the impact of  splitting the samples in two \emph{balanced}
sets rather  than splitting the  samples classified  as positives from  the ones
classified as negatives.  As shown in  the last row of Table~\ref{tab:DNA}, this
gives results  very similar  to those of  Auto-Context, demonstrating  again the
advantages of splitting the samples according to the scheme we propose.

\subsection{3D Electron Microscopy Stacks~\cite{Lucchi13a}}
\label{sec:hippocampus}

\begin{figure}[t]
  \centering
  \begin{tabular}{ccc}
    \includegraphics[width=0.35\linewidth]{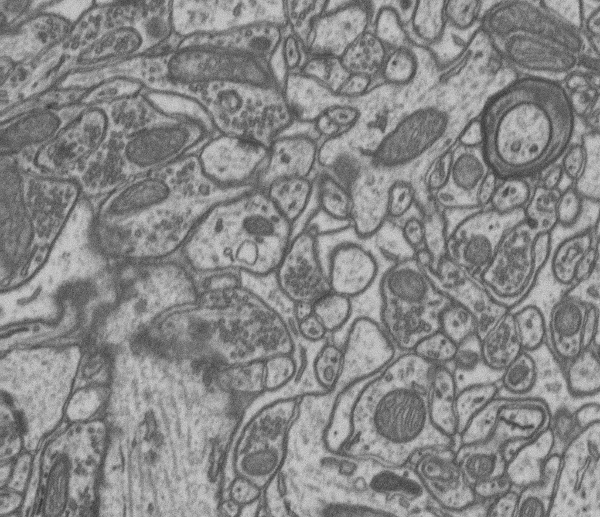} &
    \hspace{1em} &
    \includegraphics[width=0.35\linewidth]{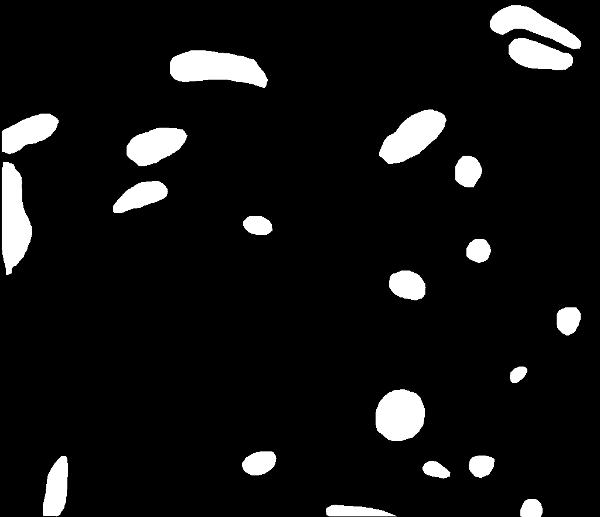} \\
    {\bf (a)} &
    \hspace{1em} &
    {\bf (b)} \\
    \includegraphics[width=0.35\linewidth]{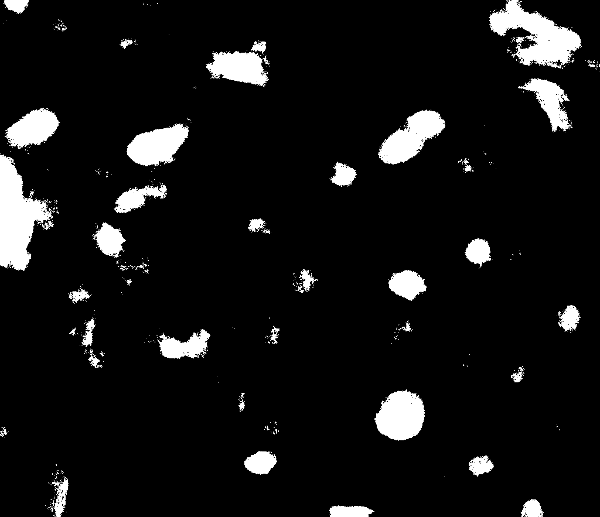} &
    \hspace{1em} &
    \includegraphics[width=0.35\linewidth]{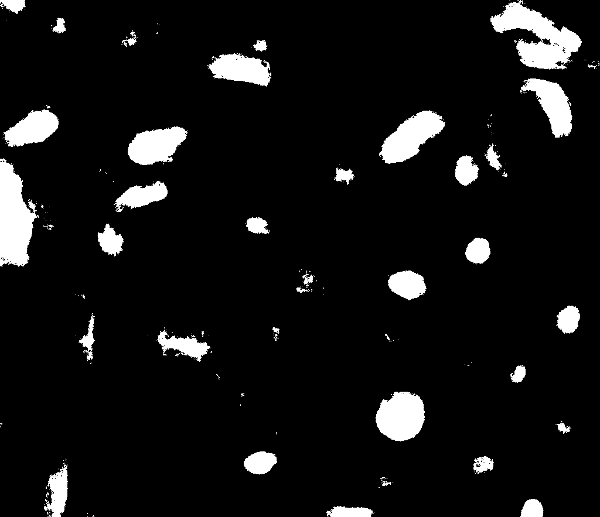} \\
    {\bf (c)} &
    \hspace{1em} &
    {\bf (d)} \\
    \includegraphics[width=0.35\linewidth]{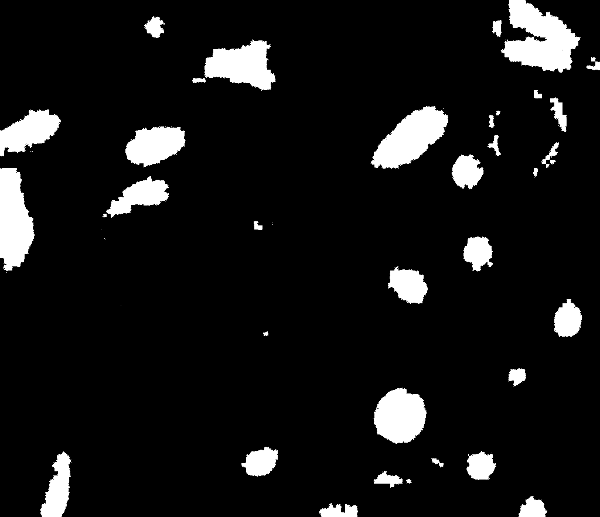} &
    \hspace{1em} &
    \includegraphics[width=0.35\linewidth]{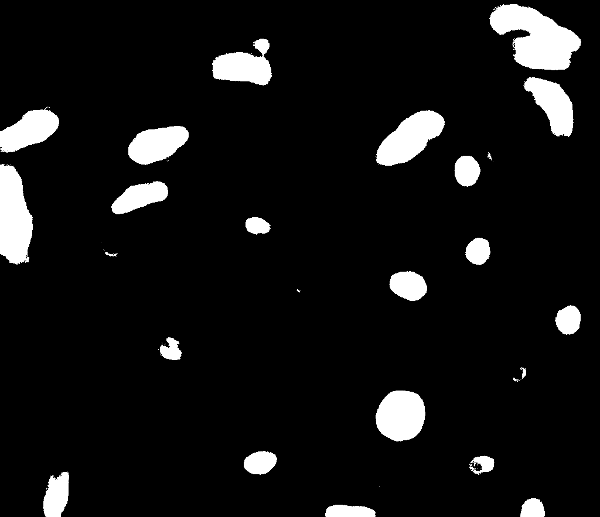} \\
    {\bf (e)} &
    \hspace{1em} &
    {\bf (f)} \\
  \end{tabular}
  \caption{Segmentation results for a test image from the CA1
    Hippocampus dataset~\cite{Lucchi13a}.
    The image has been randomly selected, but taken away from the
    stack's borders to avoid penalizing the method of~\cite{Lucchi13a}
    which otherwise could not exploit neighboring slices to improve
    its result.
    The thresholds used to obtain the segmentations from the score
    images are given by the values that maximize the VOC score for the
    given image.
    {\bf(a)} Original test image.
    {\bf(b)} Ground-truth outlining the mitochondria in the selected image.
    {\bf(c)} Segmentation obtained by KernelBoost. The VOC score on this image is 0.567.
    {\bf(d)} Segmentation obtained by KernelBoost + pooling + clustering. The VOC score on
    this image is 0.632.
    {\bf(e)} Segmentation obtained by the approach
    of~\cite{Lucchi13a}. The VOC score on this image is 0.679.
    {\bf(f)} Segmentation obtained by our approach. The VOC score on
    this image is 0.719.
  }
  \label{fig:hippocampus}
\end{figure}

Our third dataset  is  composed by two 3D  Electron Microscopy~(EM)
stacks of  the CA1 hippocampus  region of a rodent  brain~\cite{Lucchi13a}.  The
training volume  contains 165 slices with  size $1024 \times 653$  pixels, while
the test volume  has the same number  of slices but with size  $1024 \times 883$
pixels.  A sample slice, along  with its corresponding ground-truth, is depicted
by Fig.~\ref{fig:hippocampus}. The goal is to segment the mitochondria.

As  Table~\ref{tab:mitochondria}  shows,   we  outperform  the  state-of-the-art
algorithm~\cite{Lucchi13a},  even  though  we  operate on  2D  slices  only  for
simplicity, without enforcing any  consistency between consecutive slices, while
the approach of~\cite{Lucchi13a} relies on a 3D CRF.

\begin{table}[t]
  \caption{   Results   for   the  CA1   Hippocampus   dataset~\cite{Lucchi13a}.
    The values have been
    computed  by ignoring  a 1-pixel  width  region around  the
    mitochondria to account for imprecisions in the manual segmentations. See
    Section~\ref{sec:hippocampus} for more details.}
  \label{tab:mitochondria}
  \centering
  \begin{tabular}{@{}lcc@{}}
    \toprule[0.1em]
    \textbf{Method} & {\bf VOC} & {\bf F-measure} \\
    \midrule
    Lucchi~\emph{et al.}~\cite{Lucchi13a}  & 0.722         & 0.839 \\
    \KB                   & 0.625         & 0.769 \\
    \KB + pooling only    & 0.649         & 0.787 \\
    \IKB                                   & 0.711         & 0.831 \\
    {\it our approach (on \IKB)}    & {\bf 0.776}   & {\bf 0.874} \\
    \bottomrule[0.1em]
  \end{tabular}
\end{table}

\begin{figure*}[t]
  \centering
  \begin{tabular}{ccc}
    \includegraphics[width=0.38\linewidth]{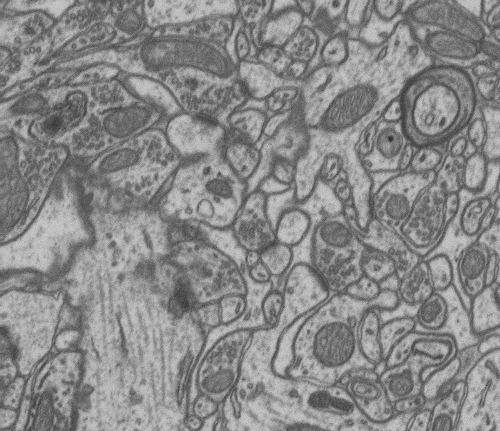} &
    \hspace{1em} &
    \includegraphics[width=0.38\linewidth]{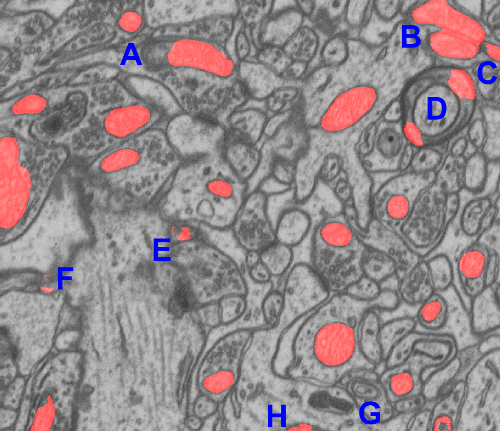} \\
    {\bf (a)} &
    \hspace{1em} &
    {\bf (b)} \\
  \end{tabular}
  \caption{Analysis of the major segmentation mistakes made by our
    algorithm on a test image from the CA1 Hippocampus
    dataset~\cite{Lucchi13a}.
    {\bf(a)} Original image.
    {\bf(b)} Original image with the segmentation we have obtained
    overlayed. We have selected this image because the
    segmentation presents several mistakes, affecting
    significantly the final score.
    (A) The strong background difference in the mitochondrion
    makes our algorithm prematurely end the segmentation.
    (B) The two mitochondria are very close, and the texture
    between them mislead our algorithm.
    (C) {\it Error in the ground-truth}: the component identified
    by our algorithm is indeed a mitochondrion.
    (D) Two shadowed parts of the image are incorrectly
    identified as mitochondria, probably due to the borders
    on both sides (such structures are not present in the
    training data).
    (E,F) Minor mistakes due to shadows in the image.
    (G) {\it Error in the ground-truth}: a component is
    incorrectly identified as a mitochondrion, while it is
    not.
    (H) Dark element on the image's border incorrectly
    identified as a mitochondrion.
    Best viewed in color.
  }
  \label{fig:hippocampus_errors}
\end{figure*}

From the table, it is also possible  to appreciate the contribution given by the
pooling and sample selection  steps as explained in Section~\ref{sec:KB_and_IKB}
with respect to the original \KB method.
A visual analysis of the segmentation mistakes is given in
Fig.~\ref{fig:hippocampus_errors}.



Furthermore, we have randomly chosen a small subset of this dataset ---
9 training and 6 test images --- to
perform an in-depth analysis of the performance of the different
KernelBoost components listed in Section~\ref{sec:KB_and_IKB}.
The results are presented in Table~\ref{tab:hipp_ikb}.

\begin{table}[t]
  \caption{Comparison of the effectiveness of the different
    KernelBoost components on a subset of the CA1 Hippocampus
    dataset~\cite{Lucchi13a}.
  }
  \label{tab:hipp_ikb}
  \centering
  \begin{tabular}{@{}lc@{}}
    \toprule[0.1em]
    \textbf{Method} & {\bf VOC} \\
    \midrule
    KB                                            & 0.592 \\
    KB + pooling                                  & 0.638 \\
    KB + clustering                               & 0.599 \\
    KB + pooling + clustering                     & 0.638 \\
    KB + Ilastik features                         & 0.667 \\
    KB + Ilastik features + pooling               & 0.673 \\
    KB + superpixel pooling                       & 0.653 \\
    KB + superpixel pooling + superpixel features & 0.660 \\
    KB + Ilastik features + superpixel pooling    & 0.683 \\
    \bottomrule[0.1em]
  \end{tabular}
\end{table}

In the case of this subset clustering does not seem to be that
relevant. This might be due to the size of the dataset, as
categorizing the positive samples might overly reduce the pool of samples
used for the filter learning step.
Pairing the input image channel with Ilastik features seems to be an
interesting opportunity, as this allows to achieve very good
performance. However, adding a pooling step on top of this does not
increase significantly the performance. This can be explained by the
fact that the smoothed images that are included in the Ilastik
features can partially compensate for the lack of a pooling step.
Superpixel pooling represents an effective strategy to improve the
performance of the system, although computationally expensive.
Adding on the top of it superpixel features does not significantly improve
the final result.
Finally, the pair superpixel pooling/Ilastik features performed best,
achieving a VOC score that is almost 10\% higher than the original
KernelBoost score.

From the insights given by these experiments, we performed extensive
tests on the architectures presented in Section~\ref{sec:KB_and_IKB}.
The results are given in Table~\ref{tab:hipp_arch}.

\begin{table}[t]
  \caption{Comparison of the architectures of
    Section~\ref{sec:KB_and_IKB} on the CA1 Hippocampus
    dataset~\cite{Lucchi13a}.
    The base result is given by KernelBoost with Ilastik features, and
    all the subsequent processing starts from the score maps this
    method produces.
    Although very effective, we have skipped superpixel pooling because
    of its high computational costs.
  }
  \label{tab:hipp_arch}
  \centering
  \begin{tabular}{@{}lc@{}}
    \toprule[0.1em]
    \textbf{Method} & {\bf VOC} \\
    \midrule
    \emph{\KB + Ilastik features}                                    & 0.654 \\
    Auto-Context                                                      & 0.701 \\
    Auto-Context + pooling (on image channel only)                    & 0.704 \\
    Auto-Context + pooling (on all channels)                          & 0.714 \\
    Auto-Context + pooling (on image channel only) + fake3D           & 0.735 \\
    Expanded Trees + pooling (on image channel only)                 & 0.720 \\
    Expanded Trees + pooling (on image channel only) + fake3D        & 0.746 \\
    Knotted Trees + pooling (on image channel only)                  & 0.717 \\
    Knotted Trees + pooling (on image channel only) + fake3D         & 0.743 \\
    Knotted Trees + pooling (on image channel only) + fake3D + Z-cut & 0.762 \\
    \bottomrule[0.1em]
  \end{tabular}
\end{table}

The results show that pooling on all the channels --- thus including
the score images from the previous iterations --- slightly improves the
segmentation results. A more significative contribution comes, as
expected, from the fake3D approach and the Z-cut.
In this experiment the Expanded Trees performed 0.3\% better than
Knotted Trees --- a negligible difference that shows that in
presence of enough training samples the two architectures are
equivalent, while the experiments on the Jurkat dataset show the
superiority of Knotted Trees when few training samples are available.

\subsection{2D Electron Microscopy Cell dataset}
\begin{figure}[t]
  \centering
  \begin{tabular}{ccc}
    \includegraphics[width=0.45\columnwidth]{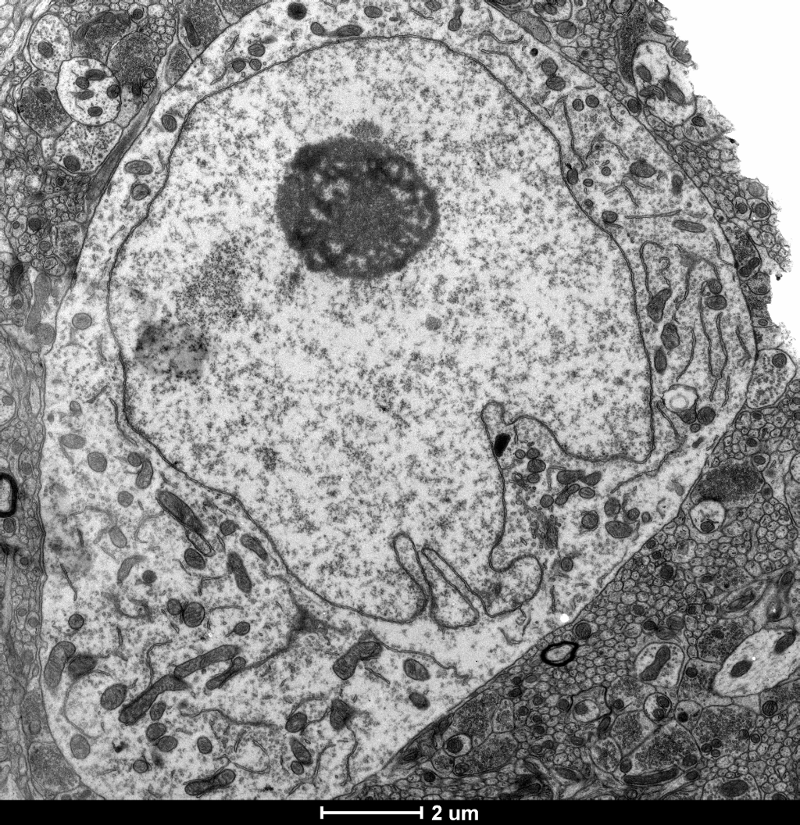} &
    \hspace{1em} &
    \includegraphics[width=0.45\columnwidth]{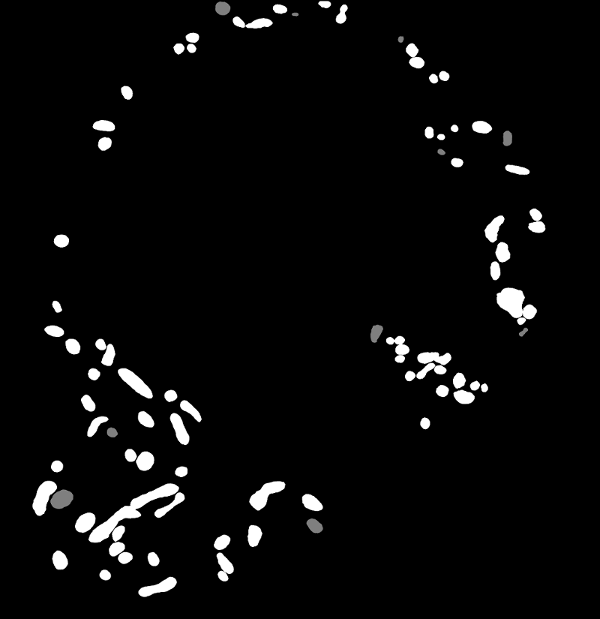}  \\
    {\bf (a)} &
    \hspace{1em} &
    {\bf (b)} \\
  \end{tabular}
  \caption{Example test image from the 2D Electron Microscopy Cell dataset,
    along with the corresponding ground-truth. Gray elements in the
    ground-truth indicate parts whose label can not be decided with
    certainty by a human expert.}
  \label{fig:BG}
\end{figure}

The last dataset is composed by 2D EM images of cells. The images are
high-dimensional (4096 $\times$ 4224), and we randomly selected 8
images for training and 4 for testing.
The goal is to segment the mitochondria inside the cell, when
masks excluding the outside of the cell and the nuclei are provided.
An example image is given in Fig.~\ref{fig:BG}, along with the
corresponding ground-truth.
Without additional information some structures appearing in the images
are not clearly categorizable by a human expert.
We have therefore introduced a third class, depicted in gray in the
ground-truth, to account for these elements.
This makes the problem more complex, but allows the final user to
individually select whether these elements have to be included in the
final result or not.

\begin{figure}[t]
  \centering
  \begin{tabular}{ccc}
    \includegraphics[width=0.45\columnwidth]{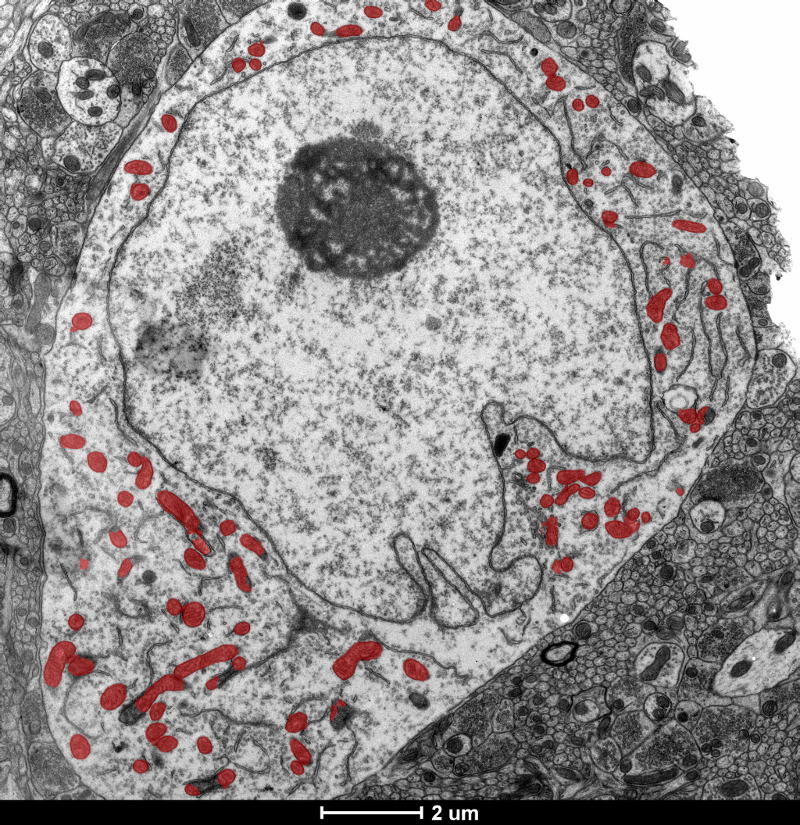} &
    \hspace{1em} &
    \includegraphics[width=0.45\columnwidth]{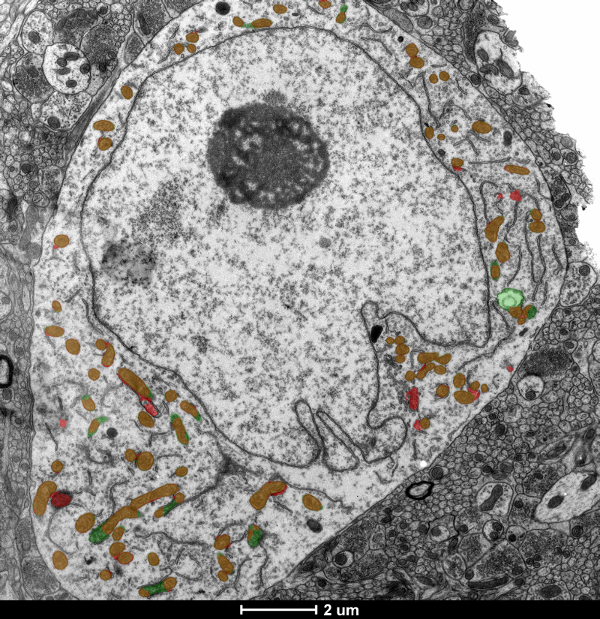}  \\
    {\bf (a)} &
    \hspace{1em} &
    {\bf (b)} \\
  \end{tabular}
  \caption{Segmentation of the image in Fig.~\ref{fig:BG}. {\bf(a)} In
  red, segmented mitochondria overlayed to the input image. {\bf(b)}
  Overlay with both the segmentation (red) and the ground-truth
  (green), which outlines the system's mistakes.}
  \label{fig:BG_res}
\end{figure}

We have experimented with multiple scales, to deal with the different
size of the mitochondria present in the images.
Also, we have used as additional channel the membranes extracted by
the approach of~\cite{Sironi14}. Note that, fed with the appropriate
ground-truth, KernelBoost is capable of extracting membranes with
an accuracy close to that of~\cite{Sironi14}, making the approach
self-contained.
From the results, presented in Table~\ref{tab:cell2D}, we can deduce
that, while both multiple scales and
membranes contribute to the final result, they work best when combined
together.
Auto-Context, preferred over Knotted Trees for its speed, significantly
contributes to the final segmentation when few weak learners are used
in the original KernelBoost classifier, while its impact is less
relevant when a large number of weak learners is used.
A visual example of the quality of the final segmentation is given in
Fig.~\ref{fig:BG_res}

\begin{table}[t]
  \caption{Evaluation of different KernelBoost components on the 2D
    Electron Microscopy Cell dataset.
  }
  \label{tab:cell2D}
  \centering
  \begin{tabular}{@{}lc@{}}
    \toprule[0.1em]
    \textbf{Method} & {\bf VOC} \\
    \midrule
    KB + multiscale, 200 weak learners                                & 0.659 \\
    KB + membranes, 200 weak learners                                 & 0.706 \\
    KB + multiscale + membranes, 200 weak learners                    & 0.762 \\
    Auto-Context (on KB + multiscale + membranes, 200 weak learners)     & 0.799 \\
    KB + multiscale + membranes, 1000 weak learners                   & 0.790 \\
    Auto-Context on (KB + multiscale + membranes, 1000 weak learners)    & 0.805 \\
    \bottomrule[0.1em]
  \end{tabular}
\end{table}

\section{Conclusion}
We have introduced a novel approach to exploiting context in a way to
capture complex interactions between neighboring image pixels.
Our method outperforms current state-of-the-art segmentation
algorithms, obtaining results which exhibit accurate boundaries
between the different image regions.

In future work we plan to operate directly on 3D data, a natural
extension of our approach, to exploit context across slices in image
stacks, and to consider different imaging types, such as Magnetic Resonance
Imaging (MRI) scans.


{\small
\bibliographystyle{ieee}
\bibliography{short,learning,vision,biomed}
}

\end{document}